\documentclass[sigconf]{acmart}

\usepackage{graphicx}
\usepackage{amsmath}
\usepackage{multirow}
\usepackage{enumitem}
\AtBeginDocument{%
  }

\usepackage{xcolor}
\usepackage{hyperref}

\copyrightyear{2026}
\acmYear{2026}
\setcopyright{cc}
\setcctype{by}
\acmConference[SIGGRAPH Conference Papers '26]{Special Interest Group on Computer Graphics and Interactive Techniques Conference Conference Papers}{July 19--23, 2026}{Los Angeles, CA, USA}
\acmBooktitle{Special Interest Group on Computer Graphics and Interactive Techniques Conference Conference Papers (SIGGRAPH Conference Papers '26), July 19--23, 2026, Los Angeles, CA, USA}
\acmDOI{10.1145/3799902.3811088}
\acmISBN{979-8-4007-2554-8/2026/07}

\acmISBN{978-1-4503-XXXX-X/2018/06}

\acmSubmissionID{466}

\def\ourMethod{UCM}

\begin{document}

\title[Unified Modeling of Camera Control and Memory for World Models]{{\ourMethod}: Unified Modeling of Camera Control and Memory with Time-aware Positional Encoding Warping for World Models}

\author{Tian-Xing Xu}
\authornote{Co-first authors.}
\email{xutx21@mails.tsinghua.edu.cn}
\affiliation{%
  \institution{Tsinghua University}
  \country{China}
}

\author{Zi-Xuan Wang}
\authornotemark[1]
\email{wangzixu21@mails.tsinghua.edu.cn}
\affiliation{%
  \institution{Tsinghua University}
  \country{China}
}

\author{Guangyuan Wang}
\authornotemark[1]
\authornote{Project leaders.}
\email{yixuan.wgy@alibaba-inc.com}
\affiliation{%
  \institution{Tongyi Lab, Alibaba}
  \country{China}
}

\author{Li Hu}
\authornotemark[2]
\email{hooks.hl@alibaba-inc.com}
\affiliation{%
  \institution{Tongyi Lab, Alibaba}
  \country{China}
}

\author{Zhongyi Zhang}
\email{ericzhang@mail.ustc.edu.cn}
\affiliation{%
  \institution{University of Science and Technology of China}
  \country{China}
}

\author{Peng Zhang}
\email{futian.zp@alibaba-inc.com}
\affiliation{%
  \institution{Tongyi Lab, Alibaba}
  \country{China}
}

\author{Bang Zhang}
\authornote{Corresponding authors.}
\email{bangzhang@gmail.com}
\affiliation{%
  \institution{Tongyi Lab, Alibaba}
  \country{China}
}

\author{Song-Hai Zhang}
\authornotemark[3]
\email{shz@tsinghua.edu.cn}
\affiliation{%
  \institution{Tsinghua University}
  \country{China}
}

\renewcommand{\shortauthors}{Tian-Xing et al.}

\begin{abstract}
  World models based on video generation demonstrate remarkable potential for simulating interactive environments yet suffer from persistent difficulties in two key areas: maintaining long-term content consistency when scenes are revisited and enabling precise camera control from user-specified inputs. Existing methods based on explicit 3D reconstruction often compromise flexibility in unbounded scenarios and struggle to preserve fine-grained structures. Alternative methods rely directly on previously generated frames without establishing explicit spatial correspondence, thereby limiting controllability and consistency. To address these limitations,  we present \textbf{\ourMethod}, a novel framework for unified modeling of long-term memory and precise camera control via a time-aware positional encoding warping mechanism. To reduce computational overhead, we design an efficient dual-stream diffusion transformer for high-fidelity generation. Moreover, we introduce a scalable data curation strategy that utilizes point-cloud-based rendering to simulate scene revisiting, enabling training on over 500K monocular videos. Extensive experiments on real-world and synthetic benchmarks demonstrate that \textbf{\ourMethod} significantly outperforms state-of-the-art methods on long-term scene consistency, while achieving precise camera controllability in high-fidelity video generation. Our code is released at {\color{blue}\url{https://humanaigc.github.io/ucm-webpage/}}.
\end{abstract}


\begin{teaserfigure}
  \includegraphics[width=\textwidth]{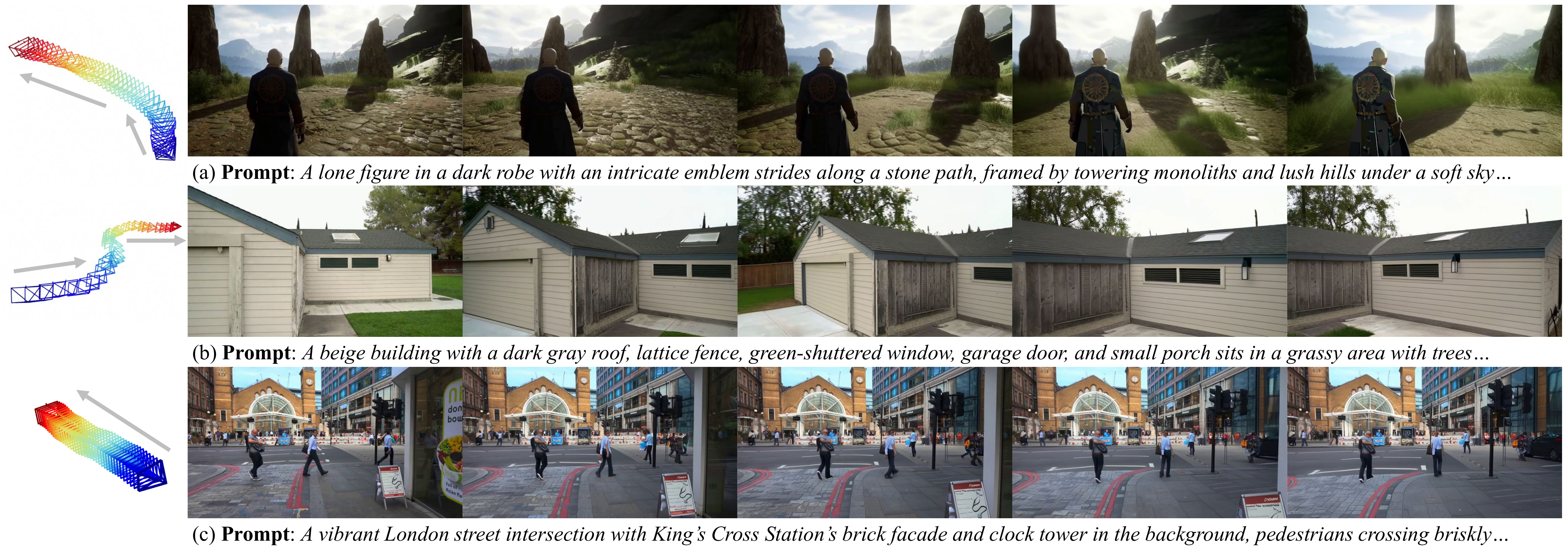}
  \caption{Visual results of our proposed {\ourMethod}. \textmd{Given a reference image, a user-specified camera trajectory and a prompt, {\ourMethod} enables camera-controlled, long-term consistent world generation via time-aware positional encoding warping.}}
  \label{fig:teaser}
\end{teaserfigure}


\maketitle

\section{Introduction}

World models~\cite{bar2025navigation,decart2024oasis,alonso2024diffusion,parker2024genie,valevski2024diffusion,authors9genesis,zhu2024sora,che2024gamegen,guo2025genesis,liu2025towards} have drawn increasing attention for their capability to simulate realistic environments in response to user inputs, serving as a fundamental pillar for diverse interactive applications, ranging from simulation~\cite{parker2024genie,zhu2024sora}, autonomous driving~\cite{guo2025genesis,liu2025towards} and robotics~\cite{authors9genesis} to game engines~\cite{valevski2024diffusion,che2024gamegen}. Recent advances~\cite{parker2024genie,yu2025context,wu2025video,gao2025longvie,liu2025worldmirror} in video-generation-based world models have substantially advanced this domain, enabling high-fidelity generation of potential future scenarios through training on large-scale real-world videos. Within this paradigm, adapting powerful video generation models for world simulation confronts two core challenges: 1) maintaining long-term content consistency and 2) achieving precise user-guided camera control. Although contemporary methods~\cite{huang2025self,zhang2025packing} ensure frame-to-frame temporal coherence, they frequently fail to maintain consistency when revisiting previously observed scenes—a limitation that is often attributed to the finite context window of temporal conditioning \cite{yu2025context,wu2025video}. Furthermore, integrating precise camera control into video generation models remains challenging, primarily due to the inherent viewpoint diversity present in open-world videos.

To address these problems, inspired by ViewCrafter~\cite{yu2024viewcrafter}, previous method~\cite{wu2025video} employs explicit 3D scene reconstruction to preserve long-term geometry and incorporate viewpoint information. It aggregates 3D point clouds estimated from all historical frames through truncated signed distance
function (TSDF) fusion~\cite{andy2017tsdf}, subsequently rendering these points from target views to condition new frame generation. However, reliance on explicit 3D representations often compromises flexibility in large-scale, unbounded scenes and can lead to loss of detail, particularly for fine-grained structures.

Another pipeline conditions future video generation directly on previously generated frames, typically by concatenating them along the temporal axis. For camera controllability and inter-frame correspondence modeling, these methods encode either raw camera parameters~\cite{yu2025context} or Pl\"{u}cker embeddings~\cite{li2025vmem,xiao2025worldmem} via a learnable camera encoder, and then inject them into the feature sequence for camera-controlled generation. Despite promising results, such methods rely on implicitly learned 3D priors—derived solely from 2D posed frames—to capture cross-view correspondences. This reliance on implicit priors impedes precise camera control and weakens spatial correspondence, ultimately resulting in content inconsistencies.

In this paper, we propose {\ourMethod}, a novel framework for unified modeling of precise camera control and long-term memory via time-aware positional encoding warping for world models. We build {\ourMethod} upon a diffusion transformer (DiT)-based video generation model, which represents videos as visual tokens augmented with 3D positional encodings (PEs) for spatio-temporal information. To condition subsequent generation, following prior works~\cite{yu2025context,li2025vmem,xiao2025worldmem}, we concatenate tokens from the reference image and historical frames as conditions. We argue that camera-controllable generation involves modeling spatial transformations between the initial and subsequent frames, while long-term memory demands spatial alignment between historical and future frames — both rely on spatio-temporal token-wise correspondences, which can be jointly addressed by our warping operation. Inspired by PE-Field~\cite{bai2025positional}, we reassign 3D PEs of conditional tokens via time-aware geometry-grounded warping, thereby providing robust, explicit spatio-temporal token correspondences for camera control and memory injection. Notably, concatenating conditional tokens extends the input sequence length. Since 3D self-attention within DiTs has quadratic complexity with respect to sequence length, this incurs considerable computational overhead. We thus present an efficient dual-stream DiT tailored for conditional generation with minimal computational cost. Additionally, a key training challenge is the scarcity of large-scale video datasets featuring long-term scene revisits from different viewpoints. To address this, we implement a scalable data curation strategy with point-cloud-based rendering to simulate scene revisiting, leveraging over 500K monocular videos across diverse scenarios for training and enhancing generalizability to open-world environments.

To evaluate {\ourMethod}, we collect open-source videos from Tanks \& Temples~\cite{knapitsch2017tanks}, RealEstate10K~\cite{zhou2018stereo}, Context-as-Memory~\cite{yu2025context}, DL3DV~\cite{ling2024dl3dv} and MiraData~\cite{ju2024miradata}, covering diverse environments and styles. Experiments show that {\ourMethod} significantly outperforms existing methods in terms of visual quality and long-term consistency upon scene revisiting, achieving state-of-the-art camera controllability. Our contributions are summarized as follows:
\begin{itemize}[leftmargin=*]
\item We introduce a novel time-aware positional encoding warping mechanism into world models, enabling the unified modeling of precise camera control and long-term scene consistency via establishing explicit spatio-temporal token correspondences.
\item We present an efficient dual-stream video diffusion model for high-fidelity generation with minimal computational overhead. 
\item We employ a simple yet effective data curation strategy to simulate long-term scene revisiting, which enables training on large-scale monocular videos and improves generalization. 
\end{itemize}

\section{Related Works}

\paragraph{Video generation models.} The recent scaling of video datasets has substantially advanced the capabilities of video generation models, such as Sora~\cite{zhu2024sora}, Seedance~\cite{gao2025seedance} and HY Video~\cite{kong2024hunyuanvideo}. Full-sequence diffusion models~\cite{gao2025seedance,kong2024hunyuanvideo,yang2024cogvideox,wan2025wan} have emerged as a predominant paradigm due to their high-quality generation. However, GPU memory constraints limit the length of generated videos, and these models generally lack scene consistency across multiple, distinct video clips. Alternative architectures, such as auto-regressive models~\cite{huang2025self,chen2024diffusion,zhang2025packing,song2025history,gu2025long}, generate new frames conditioned on preceding outputs, thereby achieving video generation of considerable length, but they are similarly constrained by a finite temporal context window, lacking long-term memory capability.

\paragraph{Memory for long-term video generation.} Many demos~\cite{song2025history,decart2024oasis,kanervisto2025world} exhibit gradual scene drift due to the limited length of the context window. To preserve long-term geometry, previous work~\cite{wu2025video} utilizes a 3D reconstruction model to estimate explicit 3D representations like point clouds from previously generated frames. These 3D representations are aggregated via TSDF fusion~\cite{andy2017tsdf} to condition subsequent clip generation. However, such explicit 3D representations often lack flexibility in large, unbounded scenes and suffer from loss of details for fine-grained structures. Other methods~\cite{li2025vmem,yu2025context,xiao2025worldmem} condition generation directly on retrieved historical frames, using metrics like view frustum similarity \cite{yu2025context,xiao2025worldmem} or 3D surfel splatting \cite{li2025vmem}. These methods rely on implicit 3D priors learned during training to model inter-frame relationships, which impedes robust long-term scene coherence across diverse scenarios.

\paragraph{Camera controlled video generation.}  Enabling video generation conditioned on explicit camera trajectories remains a central challenge for world models. One line of work employs explicit 3D representations derived from an initial frame to guide image-to-video generation, utilizing techniques such as point cloud rendering~\cite{cao2025uni3c,feng2024i2vcontrol,li2025realcam,ma2025you,you2024nvs,yu2025trajectorycrafter,yu2024viewcrafter,zhai2025stargen}, tracking~\cite{gu2025diffusion} or optical flow~\cite{burgert2025go}. Alternatively, other approaches incorporate additional trainable modules into existing video diffusion models to learn the implicit frame-wise correspondence from data, conditioning on raw pose parameters~\cite{bai2025recammaster,wang2024motionctrl}, Pl\"{u}cker embeddings~\cite{bahmani2025ac3d,he2024cameractrl,he2025cameractrl} or relative camera encodings~\cite{zhang2025unified}. However, such implicit correspondence exhibits poor generalization for complex camera trajectories or excessive camera movements, failing to achieve the same precise camera controllability as 3D reconstruction-based novel view synthesis~\cite{dai2024real,wu2024recent,jing2025frnerf}.

\section{Preliminaries}

\paragraph{DiT-based video generation models.} Our method is built upon a pretrained image-to-video (I2V) generation model~\cite{wan2025wan}. This model consists of a causal spatio-temporal Variational Autoencoder~\cite{kingma2013auto} (VAE), which learns compact latent representations from high-dimensional visual data, and a latent diffusion transformer~\cite{peebles2023scalable} (DiT) that models the data distribution through iterative denoising. Each transformer block is instantiated as a sequence of 3D self-attention for modeling spatio-temporal relationships, cross-attention to integrate text-conditioned information, and a feed-forward network (FFN) for feature refinement. Following Rectified Flows~\cite{esser2024scaling}, the forward diffusion process is defined as $\mathbf{x}_t = t \mathbf{x}_1 + (1-t) \mathbf{x}_0$, where $\mathbf{x}_1$ denotes the clean latent code encoded by the causal VAE, $\mathbf{x}_0 \sim \mathcal{N}(0, I)$ denotes Gaussian noise, and the timestep $t\in [0, 1]$ is sampled from a predefined distribution. The latent transformer $u_\theta$ learns to predict the velocity field $\mathbf{v}_t = d\mathbf{x}_t/dt = \mathbf{x}_1 - \mathbf{x}_0$, which defines an ordinary differential equation (ODE), by minimizing the training objective
\begin{equation}
    \mathcal{L}(\theta) = \mathbb E_{\mathbf{x}_0, \mathbf{x}_1, t} ||u_\theta(\mathbf{x}_0,\mathbf{x}_1,t) - \mathbf{v}_t||_2^2
\end{equation}
 Here, $\theta$ denotes the learned model weights. During inference, the network iteratively transforms randomly sampled Gaussian noise into a clean latent representation, which the VAE then decodes to generate the final video. 
 
\paragraph{Positional encoding field (PE-Field).} In DiT-based image generation models, the latent code $\mathbf{x}$ is patchified and flattened into a sequence of tokens.
2D positional encodings (PEs), particularly RoPEs~\cite{su2024roformer}, are appended to each token to indicate its 2D spatial locations, primarily enforcing spatial coherence within the self-attention mechanism~\cite{bai2025positional}. Motivated by this finding, PE-Field~\cite{bai2025positional} formulates novel view synthesis (NVS) as image generation conditioned on a source image and relative camera transformations. It concatenates clean tokens from the source image with noisy target tokens along the sequence dimension, then reassigns the PEs of clean tokens according to their projected positions, derived from 3D scene reconstruction and target view transformation. Given that patch tokens are spatially coarser than pixel-wise warping, PE-Field proposes multi-level PEs for sub-patch detail modeling to improve alignment precision, which apply different heads of attention layers with warped PEs derived from different resolution grids. Additionally, PE-Field extends PEs with per-token depth values, enabling the DiT to explicitly model relative depth relationships.
\section{Method}

\begin{figure*}[h]
  \centering
  \includegraphics[width=\linewidth]{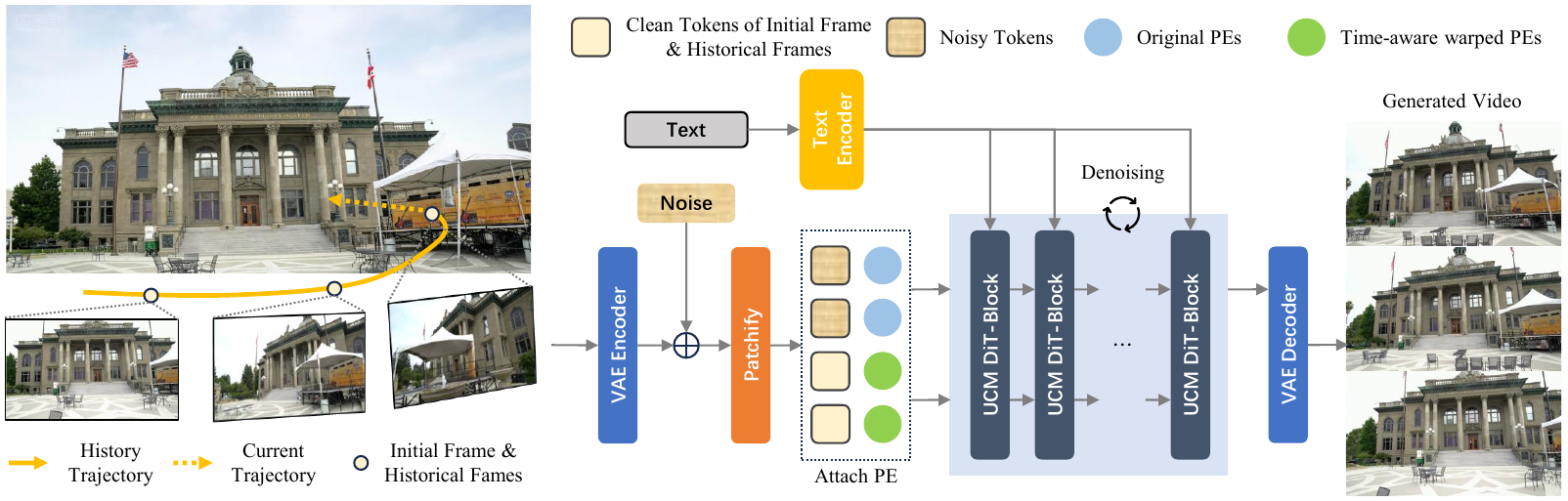}
  \caption{An overview of {\ourMethod}. Given previously generated frames and a specific camera trajectory as input, {\ourMethod} encodes the historical frames into clean tokens to condition the denoising of noisy tokens. It adopts time-aware positional encoding warping for unified modeling of camera-controllable generation and memory injection. After iterative denoising, {\ourMethod} yields a high-fidelity, scene-consistent video following the user-specified trajectory. }
  \label{fig:overview}
\end{figure*}

Starting from a reference image $I^{r} \in \mathbb{R}^{H\times W \times 3}$, we aim to leverage powerful video generation models for world simulation guided by a user-specified camera trajectory. Our method adopts a clip-by-clip generation paradigm for long-term simulation, where each clip $V=\{I_i\}_{i=1}^T\in \mathbb{R}^{T\times H\times W \times 3}$ is conditioned on either the reference image or the last frame of the preceding clip. To ensure long-term scene coherence, we follow previous approaches~\cite{yu2025context,li2025vmem,xiao2025worldmem} by retrieving the most relevant historical frames $\{I_{h_j}\}_{j=1}^M \in \mathbb R^{M\times H \times W \times 3}$ with corresponding view matrices $\{c_{h_j}\}_{j=1}^M\in \mathbb R^{M\times 4 \times 4}$, which serve as memory to condition the generation of subsequent clips. The overview of our proposed {\ourMethod} is illustrated in Fig. ~\ref{fig:overview}, which achieves the unified modeling of camera control and memory injection through time-aware positional encoding warping (Sec.~\ref{sec:pe_warping}). This warping operation establishes robust spatio-temporal token correspondences, where the conditional information is integrated via an efficient dual-stream video diffusion architecture with minimal computational overhead (Sec.~\ref{sec:light-weight}). To address the scarcity of large-scale videos featuring long-term multiple revisiting, we adopt a simple yet effective dataset curation strategy, facilitating model training on large-scale monocular video datasets (Sec.~\ref{sec:dataset}).

\subsection{Time-aware Positional Encoding Warping}
\label{sec:pe_warping}

Given retrieved memory frames $\{I_{h_j}\}_{j=1}^M$ with view matrices $\{c_{h_j}\}_{j=1}^M$ and a reference frame $I^r$ as the conditional image, our objective is to generate a high-fidelity video $V=\{I_i\}_{i=1}^T$ that adheres to a text prompt and a specific new camera trajectory. We first apply the 3D VAE and patchify operation to the memory frames, conditional image and target video for dimension compression, obtaining latent codes $\mathbf{x}_h = \{x_{h_j}\}_{j=1}^M \in \mathbb R^{M\times \tilde{H} \times \tilde{W} \times D}$, $x^r \in \mathbb{R}^{\tilde{H}\times \tilde{W} \times D}$ and $\mathbf{x} = \{x_i\}_{i=1}^N \in \mathbb R^{N\times \tilde{H}\times \tilde{W} \times D}$, respectively. Supposing $s$ and $r$ denote the spatial and temporal compression ratios, the shape of the latent codes satisfies $\tilde{H} = H/s$, $\tilde{W} = W/s$, and $N = (T+r-1)/r$. To achieve temporal alignment between the camera trajectory and the latent sequence, we assume the view transformation matrices change uniformly within $r$ continuous frames, applying average pooling to the input trajectory to obtain $\mathbf{c} = \{c_i\}_{i=1}^{N} \in \mathbb R^{N\times 4\times 4}$. These latent codes are then flattened into a sequence of tokens and processed by DiT blocks, which are adapted to learn the conditional distribution
\begin{equation}
    \mathbf{x} \sim p(\mathbf{x} | \mathbf{x}_h, x^r, \mathbf{c}, \mathbf{c}_h, w)
\end{equation}
where $w$ represents the user-provided text condition. Existing I2V models~\cite{wan2025wan,kong2024hunyuanvideo} typically treat the reference image as the first frame to guide synthesis. For notational simplicity, we consider the reference image as a special historical frame $I_{h_0}=I^r$ with an associated camera pose $c_{h_0}=c_1$, forming $\overline{\mathbf{x}}_h = \{x_{h_j}\}_{j=0}^M$ and $\overline{\mathbf{c}}_h = \{c_{h_j}\}_{j=0}^M$.

To model the relationship between the historical frames and the target views, previous works concatenate these conditional codes $\overline{\mathbf{x}}_h$ to the noisy codes $\mathbf{x}_t$ along the temporal axis before flattening, employing an auxiliary camera encoder to inject raw camera parameters ~\cite{yu2025context} or Pl\"{u}cker embeddings ~\cite{xiao2025worldmem,li2025vmem} into the generation process. These methods establish only frame-level viewpoint correspondence, relying on implicit 3D priors learned during training, thereby limiting their ability to track complex camera trajectories and maintain long-term consistency. To address this, we introduce time-aware PE warping, inspired by PE-Field~\cite{bai2025positional}, for unified modeling of camera control and memory for world models. Specifically, existing DiT-based methods apply 3D PEs to each visual token to capture inter-token relationships, which are obtained from their 3D coordinate $(t, u, v)$. We first estimate a sequence of depth maps $\{D_{h_j}\}_{j=0}^M \in \mathbb{R}^{(M+1)\times H\times W}$ for memory frames and reference image via a streaming depth estimation method~\cite{lan2025stream3r}, then lift them into point clouds $\{\mathcal{P}_{h_j}\}_{j=0}^M$ via the given view matrices $\overline{\mathbf{c}}_h$ through inverse perspective projection $\phi^{-1}$
\begin{equation}
    \mathcal{P}_{h_j} = \phi^{-1} (D_{h_j}, c_{h_j})
\end{equation}
With the point cloud $\mathcal{P}_{h_j}$, we can project it into the camera coordinate system of $i$-th target frame using the view transformation matrices $c_i$, obtaining warped image coordinate maps for each pixel of the historical image $I_{h_j}$
\begin{equation}
    \left [U^{h_j}_i, V^{h_j}_i\right] = \phi(\mathcal{P}_{h_j}, c_i)
\end{equation}
where $U^{h_j}_i, V^{h_j}_i \in \mathbb R^{H\times W}$. These coordinate maps are downsampled to match the spatial resolution of the latent codes $\overline{\mathbf{x}}_h$ and augmented with the temporal index $i$ to form the time-aware warped positional encoding $W_i^{h_j} = [i, U_{i}^{h_j}, V_{i}^{h_j}]$ for each conditional code $x_{h_j}$.

A key consideration is determining the target viewpoints for warping each conditional token $x_{h_j}$, because exhaustively warping to all $N$ viewpoints would introduce unacceptable computational complexity. Thus, for frame-level camera control, we replicate the visual code $x_{h_0}$ of the reference image $N$ times, warping their positional encodings to each target viewpoint $c_i$. For memory-guided generation, each historical frame $I_{h_j}$ is projected only to its most relevant viewpoint $k_j$, obtaining the final conditional token sequence with time-aware warped PEs as
\begin{equation}
    \left\{\left(x_{h_0}, W^{h_0}_i\right)\right\}_{i=1}^{N} \bigcup \left\{\left(x_{h_j}, W^{h_j}_{k_j} \right) \right\}_{j=1}^{M}
    \label{eq:cond}
\end{equation}
These conditional tokens with time-aware warped PEs are then concatenated with the noisy tokens and fed into DiT blocks to guide camera-controlled, scene-coherent video generation. Following PE-Field~\cite{bai2025positional}, we employ multi-level PEs to enhance sub-patch alignment precision. Unlike PE-Field, we do not explicitly incorporate depth values into the PEs, as the temporal coherence of video data enables the model to learn relative depth relationships implicitly.

\subsection{Efficient Dual-stream Video Diffusion}
\label{sec:light-weight}

\begin{figure}[h]
  \centering
  \includegraphics[width=\linewidth]{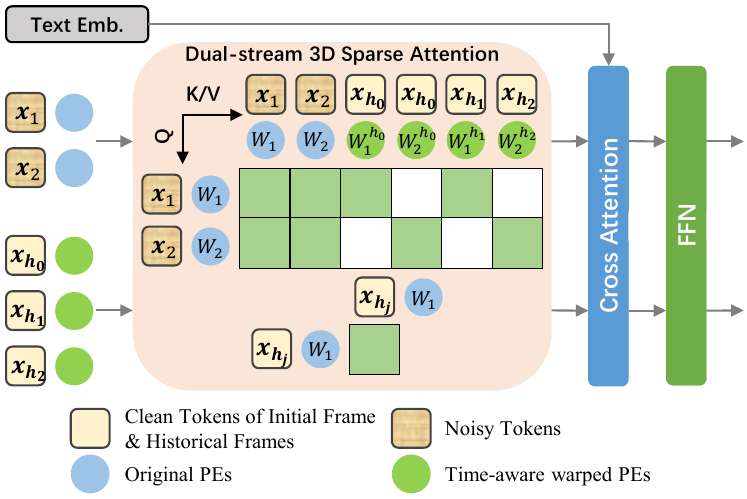}
  \caption{{\ourMethod} DiT-block. Each noisy token attends to all other noisy tokens and is guided by clean tokens via time-aware warped PEs (KV concatenation). Each clean token only attends to intra-frame peers using original PEs. This block-sparse attention mask (here, with $k_j=j$ for visualization) enables conditional generation with reduced computational cost.}
    \label{fig:light_dit}
\end{figure}

Although the time-aware warped PEs in Eq.~\ref{eq:cond} establish strong, explicit spatio-temporal correspondence between tokens, the computational overhead from the additional tokens constrains the handling of extensive memory frames. Notably, the input tokens to the DiT can be categorized into two groups: clean tokens serve as conditioning signals to guide denoising, while noisy tokens represent the generated content and require complex modeling through iterative denoising. Building on this observation, we propose an efficient dual-stream video diffusion model, composed of sequential {\ourMethod} DiT-blocks. As shown in Fig.~\ref{fig:light_dit}, each block processes visual tokens through dual-stream 3D sparse attention, followed by a cross-attention layer to inject the text prompt and a feed-forward network (FFN) for feature refinement. For each clean token from the conditional code $x_{h_j}$, we restrict these tokens to attend only to other tokens from $x_{h_j}$, while the keys and values of these tokens are concatenated with warped time-awared PEs to noisy tokens to guide content generation. For the noisy tokens, in addition to the inherited 3D full attention among noisy tokens, the strong spatio-temporal correspondence provided by time-aware PE warping enables the application of a binary attention mask. This mask forces each noisy tokens, as a query, to attend only to those clean tokens warped into the same camera views. Leveraging the block sparsity of attention, our method achieves high-fidelity video generation with precise camera control and consistent content, while incurring only minimal computational overhead.

\subsection{Data Curation}
\label{sec:dataset}

\begin{figure}[h]
  \centering
  \includegraphics[width=\linewidth]{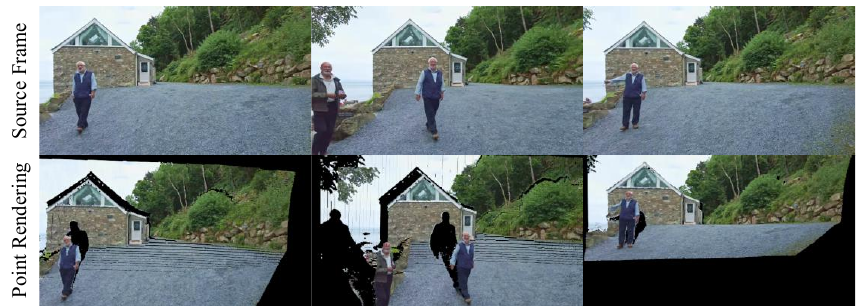}
  \caption{Simulated revisiting. We apply point cloud rendering with randomly perturbed viewpoints to simulate scene revisiting for monocular videos.}
    \label{fig:dataset}
\end{figure}

Training {\ourMethod} ideally requires long-term videos with multiple scene revisits from varying viewpoints. However, existing datasets are either collected under pure multi-view settings~\cite{ling2024dl3dv,roberts2021hypersim,dai2017scannet}, containing only static scenes without dynamic foreground objects, or suffer from limited scale and visual diversity. Alternative methods utilizing render engines like Unreal Engine 5 to synthesize multi-camera~\cite{bai2025recammaster} or long-term revisitation videos~\cite{yu2025context} often produce non-photorealistic imagery, thereby limiting world models' generalization to real-world scenarios. To overcome these limitations, we adopt a simple yet effective data curation strategy, training our model on large-scale monocular videos. Given a monocular video $V=\{I_i\}_{i=1}^T$, we leverage a 3D reconstruction model~\cite{lin2025depth} to obtain point clouds $\{\mathcal{P}_{i}\}_{i=1}^T$ and the associated camera trajectory $\{c_i\}_{i=1}^T$. To simulate scene revisits, we randomly select frames and render their point clouds $\mathcal{P}_i$ from novel viewpoints, defined by random camera offsets $\Delta c$, each yielding a rendered image $I_i'\in \mathbb R^{H\times W \times 3}$ with a binary mask $\mathcal M_i'\in \mathbb{R}^{H\times W}$ that indicates occluded/out-of-frame regions, as shown in Fig.~\ref{fig:dataset}. Since our I2V model accepts a binary mask concatenated with the latent codes as input to indicate the preserved frame, we replace it with $\mathcal M_i'$, explicitly informing the model which historical tokens reliably guide high-fidelity generation. We further warp $I_i'$ to frame $I_{i+\Delta i}$ with a random temporal shift $\Delta i$, introducing temporal geometric misalignment across different viewpoints for dynamic contents. 
Notably, though TrajectoryCrafter~\cite{yu2025trajectorycrafter} also utilizes point cloud rendering for view augmentation, it relies on double projection to produce pixel-aligned partial conditional videos. In contrast, our random temporal shift enforces static scene consistency while encouraging the model to disregard dynamic objects, whose motions violate cross-view geometric constraints. This enables our method to ignore inconsistent dynamic elements when retrieving historical cues for scene revisiting, allowing {\ourMethod} to train on large-scale monocular videos.

\section{Experiments}

\subsection{Implementation Details}

\paragraph{Training.} Our {\ourMethod} is built upon an internal I2V model finetuned from Wan2.1 1.3B T2V model~\cite{wan2025wan}, supporting 81-frame (21 latent frames) 640×352 video generation. We train on 561k monocular videos with large camera motion from Miradata~\cite{ju2024miradata}, SpatialVID~\cite{wang2025spatialvid} and Context-as-Memory~\cite{yu2025context} (801 frames each). Camera poses and point clouds are annotated via Depth Anything 3~\cite{lin2025depth}. {\ourMethod} is trained with AdamW~\cite{loshchilov2017decoupled} (learning rate $3 \times 10^{-6}$) for 30K iterations on 8 NVIDIA-A100 GPUs (batch size 8, $\sim$ 4 days).

\begin{table}[t]
  \caption{Quantitative comparison of camera controllability. We highlight the \textbf{best} and \underline{second best} entries.}
  \label{tab:cam}
  \setlength{\tabcolsep}{2pt}
  \begin{tabular}{c|c|cc|cc}
    \toprule
    \multirow{2}{*}{Method} & \multirow{2}{*}{Camera Enc.} & \multicolumn{2}{c|}{Camera Control} & \multicolumn{2}{c}{Visual Quality}  \\
    & & RotErr ($^\circ$) ↓ & TransErr ↓ & FID ↓ & FVD ↓ \\
    \midrule
    UCPE & Relative Enc. & 6.28 & 0.39 & 86.32 & 422.36\\
    C-a-M & Raw Param. & 5.45 & 0.46 & 81.99 & 377.56\\
    VMem & Pl\"{u}cker & 2.22 & 0.30 & 79.14 &362.91\\
    VWM & Point cloud & \underline{1.54} & \textbf{0.10}&\textbf{68.79}&\textbf{250.49}\\
    Ours & TPE Warping & \textbf{1.01} & \underline{0.11} & \underline{69.76} & \underline{261.27}\\
  \bottomrule
\end{tabular}
\end{table}

\begin{figure}[t]
  \centering
  \includegraphics[width=\linewidth]{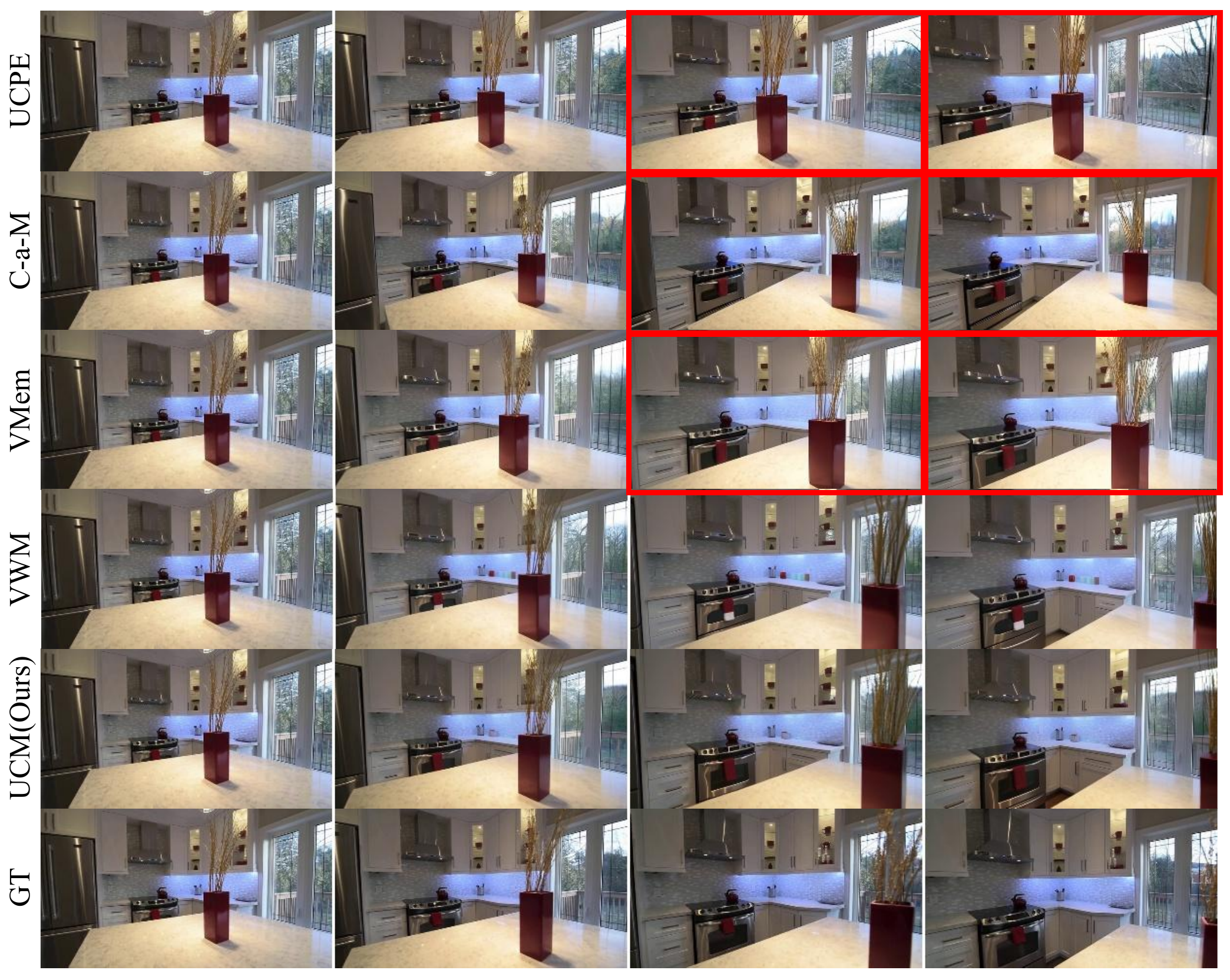}
  \caption{Visual comparison of camera controllability. We highlight imprecise camera-controlled frame generation with red boxes.}
    \label{fig:cam}
    \vspace{-1em}
\end{figure}

\paragraph{Inference.} We adopt STream3R~\cite{lan2025stream3r} to estimate depth maps for generated frames. Following previous methods~\cite{yu2025context,xiao2025worldmem}, we retrieve 20 relevant historical frames with the co-visibility of Fields of View (FoV), defined by the Intersection over Union (IoU) ratio between target and historical camera views. For each target latent frame  (except the first, conditioned on a reference image), we warp the most similar historical frame to its viewpoint. We employ Classifier-Free Guidance (CFG)~\cite{Jonathan2022classifier} for text-guided sampling, with 50 steps.

\begin{table*}[htbp]
  \caption{Quantitative evaluations for long-term memory persistency. We highlight the \textbf{best} and \underline{second best} entries.}
  \label{tab:mem}
  \setlength{\tabcolsep}{2pt}
  \resizebox{\linewidth}{!}{
  \begin{tabular}{c|cc|cc|ccc|cc|cc|ccc}
    \toprule
    \multirow{3}{*}{Method} &\multicolumn{7}{c|}{Memory Initialization} & \multicolumn{7}{c}{Cycle Trajectory}\\
      & \multicolumn{2}{c|}{Camera Control} & \multicolumn{2}{c|}{Visual Quality} &\multicolumn{3}{c|}{View Recall Consistency} & \multicolumn{2}{c|}{Camera Control} & \multicolumn{2}{c|}{Visual Quality} &\multicolumn{3}{c}{View Recall Consistency} \\
    & RotErr ($^\circ$) ↓ & TransErr ↓ & FID ↓ & FVD ↓ & SSIM ↑ & PSNR ↑ & LPIPS ↓ & RotErr ($^\circ$) ↓ & TransErr ↓ & FID ↓ & FVD ↓ & SSIM ↑ & PSNR ↑ & LPIPS ↓\\
    \midrule
     C-a-M & 10.87 & 0.39 & 115.17 & 303.34 & 0.35 & 12.44 & 0.56 & 11.50 & 0.52 & \underline{37.51} & 138.29 & 0.50 & 15.68 & 0.34\\
     VMem & 5.35 & 0.23 & \underline{110.91} & \underline{292.08} & 0.38 & 12.78 & 0.50 & 4.78 & 0.35 & 44.05 & \underline{133.95} & 0.55 & 16.35 & 0.28\\
     VWM & \underline{2.60} & \underline{0.13} & 115.11 & 301.77 & \underline{0.46} & \underline{15.06} & \underline{0.42} & \underline{3.54} & \underline{0.13} & 61.43 & 179.72 & \underline{0.68} & \underline{19.54} & \underline{0.23}\\
     Ours & \textbf{2.32} & \textbf{0.12} & \textbf{83.44} & \textbf{198.84} & \textbf{0.48} & \textbf{15.57} & \textbf{0.34} & \textbf{3.29} & \textbf{0.12} & \textbf{21.78} & \textbf{61.47} & \textbf{0.77} & \textbf{23.01} & \textbf{0.09}\\
  \bottomrule
\end{tabular}
}
\end{table*}

\begin{figure*}[h]
  \centering
  \includegraphics[width=\linewidth]{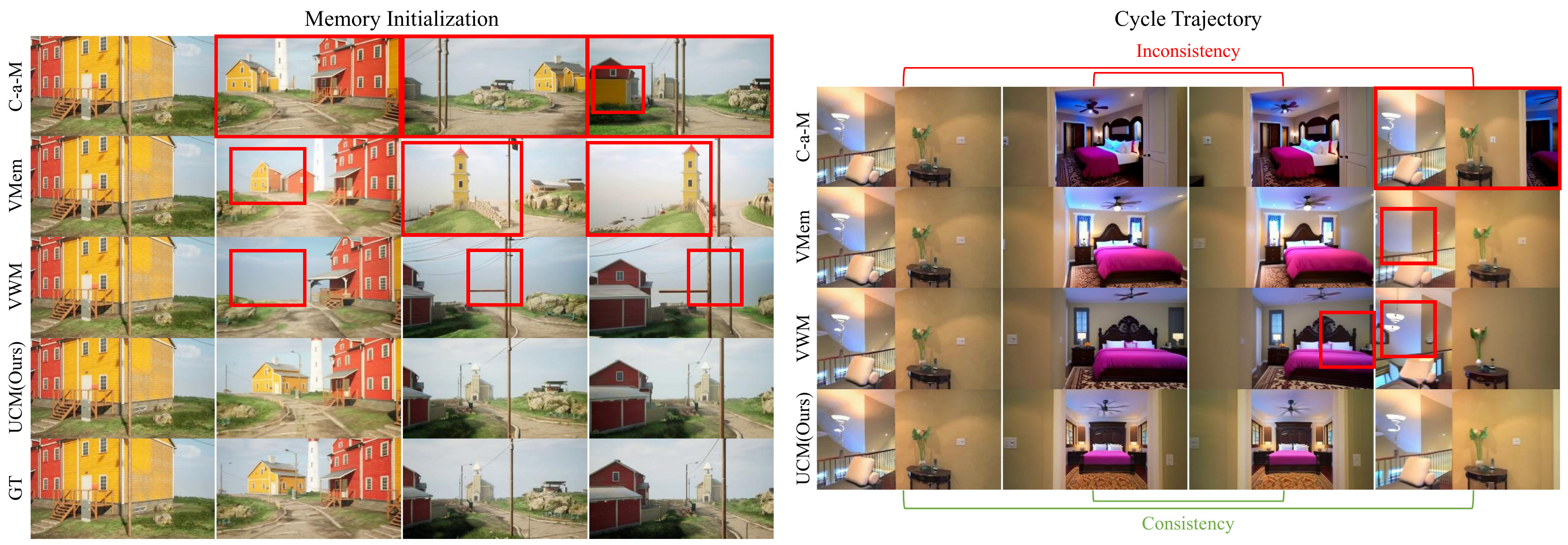}
  \caption{Visual comparison of long-term memory. Red boxes highlight obvious failure cases of camera-controlled generation or inconsistent scene generation.}
  \label{fig:main_experiment}
\end{figure*}

\paragraph{Evaluation.} We evaluate {\ourMethod} along two primary dimensions: camera controllability and long-term scene consistency. For quantitative assessment, we collect 112 diverse videos of static scenes from Realestate10K~\cite{zhou2018stereo}, Tanks-and-Temples~\cite{knapitsch2017tanks}, and a 5\% held-out subset of Context-as-Memory~\cite{yu2025context}. These videos cover diverse indoor and outdoor scenarios with both realistic and synthetic styles. For qualitative comparison, we additionally use held-out videos of dynamic scenes from MiraData~\cite{ju2024miradata} and videos of static scenes from DL3DV~\cite{ling2024dl3dv}. We use the following metrics:
\begin{itemize}[leftmargin=*]
\item {\bfseries Camera Control.} To quantify alignment between the camera trajectories of generated and ground-truth videos, we employ Depth Anything 3~\cite{lin2025depth} to extract camera poses from videos. Following CameraCtrl~\cite{he2024cameractrl}, camera poses are expressed relative to the first frame with normalized translation. We report the SO3 rotation distance (\textbf{RotErr}) and the $L_2$ translation distance (\textbf{TransErr}).
\item {\bfseries Visual Quality.} For image-level and video-level quality assessment, we calculate Fréchet Inception Distance (\textbf{FID})~\cite{heusel2017gans} and Fréchet Video Distance (\textbf{FVD})~\cite{unterthiner2018towards} between the ground truth and generated videos, respectively.
\item {\bfseries View Recall Consistency.} We employ \textbf{PSNR}~\cite{wang2002universal}, \textbf{SSIM}~\cite{wang2004image}, and \textbf{LPIPS}~\cite{zhang2018unreasonable} to measure the similarity between image pairs from identical viewpoints.
\end{itemize}

\subsection{Camera Control}

For camera controllability, we compare {\ourMethod} with representative video generation-based world models, including Context-as-Memory (C-a-M)~\cite{yu2025context} and VMem~\cite{li2025vmem}, which encode raw camera parameters and Plücker embeddings via a camera encoder for implicit 3D priors, and Video World Model (VWM)~\cite{wu2025video}, which utilizes 3D point cloud renderings as conditions. Furthermore, we also evaluate the state-of-the-art relative camera encoding method UCPE~\cite{zhang2025unified} on 81-frame videos extracted from the videos we collected. Due to inaccessible weights~\cite{yu2025context,wu2025video} or their unsuitability for I2V tasks~\cite{li2025vmem,zhang2025unified}, we reimplement these methods on the same 1.3B-parameter model as ours, trained with our data curation strategy. As shown in Tab.~\ref{tab:cam} and Fig.~\ref{fig:cam}, {\ourMethod} outperforms implicit camera-controlled methods significantly, demonstrating the effectiveness of time-aware positional encoding warping. While VWM achieves state-of-the-art performance, it is limited by 3D representation quality in unbounded scenes and fine-grained details, as discussed in Sec.~\ref{sec:memory} and shown in Fig.~\ref{fig:supp_experiment}.

\subsection{Long-term Memory}
\label{sec:memory}

We compare our method with C-a-M~\cite{yu2025context}, VMem~\cite{li2025vmem} and VWM~\cite{wu2025video} under two evaluation protocols, following previous methods~\cite{yu2025context}.
\begin{itemize}[leftmargin=*]
\item {\bfseries Memory Initialization.} For each 801-frame video, we utilize the previous consecutive 480 frames as historical frames to predict the following 321 frames. Notably, we exclude the videos from RealEstate10K~\cite{zhou2018stereo} for their short durations. The quality of the 321 generated frames is assessed through direct comparison with the ground truths. 
\item {\bfseries Cycle Trajectory.} Given the initial frame and a text prompt as conditions, we generate a long-term video that adheres to the cycle camera trajectory by making the camera return to the starting point along the same path in reverse order. For visual quality metrics, we evaluate whether newly generated frames match historical temporally symmetric generated frames.
\end{itemize}

\begin{table}[t]
  \caption{Inference speed comparison across different methods.}
  \label{tab:infer}
  \setlength{\tabcolsep}{2pt}
  \begin{tabular}{c|cc|c}
    \toprule
    Method & Gen. (s/frame) & Avg. 3D Recon. (s/frame) & Total (s/frame)  \\
    \midrule
    C-a-M & 2.86 & - & 2.86\\
    VMem & 3.31 & 0.40 & 3.71\\
    VWM & 1.71 & 0.40 & 2.11\\
    Ours & 2.40 & 0.36 & 2.76\\
  \bottomrule
\end{tabular}
\end{table}

\begin{table*}[htbp]
  \caption{Quantitative evaluations for ablation studies on sparse attention and the number of memory frames. We highlight the \textbf{best} and \underline{second best} entries, and employ the symbol {\textdagger} to specify default configurations. ``Mem'' indicates the number of retrieved memory frames, while ``Dual'' and ``Sparse'' represent dual-stream architecture and sparse attention, respectively. ``Data'' is the data curation strategy.}
  \label{tab:ablate}
  \resizebox{\linewidth}{!}{
  \begin{tabular}{cccc|cc|ccc|cc|ccc|c}
    \toprule
    
     & & &  &\multicolumn{5}{c|}{Memory Initialization} & \multicolumn{5}{c|}{Cycle Trajectory} & Generation  \\ 
     & & & & \multicolumn{2}{c|}{Visual Quality} &\multicolumn{3}{c|}{View Recall Consistency} & \multicolumn{2}{c|}{Visual Quality} &\multicolumn{3}{c|}{View Recall Consistency} & Speed\\
   Mem & Dual & Sparse & Data & FID ↓ & FVD ↓ & SSIM ↑ & PSNR ↑ & LPIPS ↓ & FID ↓ & FVD ↓ & SSIM ↑ & PSNR ↑ & LPIPS ↓ & s/frame ↓ \\
     \midrule
     2 & \checkmark & \checkmark & \checkmark & 88.24 & 226.35 & 0.47 & 15.43 & 0.36 & 27.81 & 75.25 & 0.65 & 20.14 & 0.15 & 1.72\\ 
     4 & \checkmark & \checkmark & \checkmark & 84.92 & 207.05 & 0.47 & 15.49 & 0.35 & 24.92 & 67.32 & 0.69 & 21.35 & 0.12 & 1.80 \\
     5 & \checkmark & \checkmark & \checkmark & 85.29 & 213.19 & 0.47 & 15.54 & 0.35 & 24.46 & 68.49 & 0.70 & 21.59 & 0.12 & 1.81\\
     10 & \checkmark & \checkmark & \checkmark & 84.05 & 211.89 & \underline{0.48} & 15.49 & 0.35 & 22.98 & 61.82 & 0.74 & 22.49 & 0.10& 2.03\\
     20$^\dagger$ & \checkmark & \checkmark & \checkmark  & \underline{83.44} & \underline{198.84} & \underline{0.48} & \underline{15.57} & \textbf{0.34} & 23.01 & 58.16 & \underline{0.77} & 23.57 & 0.09 & 2.40 \\
     40 & \checkmark & \checkmark & \checkmark & \textbf{80.68}& \textbf{188.41}&  \textbf{0.49}& \textbf{15.79}& \textbf{0.34}& \underline{18.76}& \underline{52.72}&\textbf{0.80} & \textbf{23.79}& \textbf{0.08}& 3.26\\
     20 &  &  & \checkmark & 85.05 & 205.43 & 0.47 & 15.32 & 0.35 & \textbf{17.13} & \textbf{37.55} & \underline{0.77} & \underline{23.76} & \textbf{0.08}&5.14\\
     20 & \checkmark &  & \checkmark & 94.72 & 270.09 & 0.47 & 14.30 & 0.37 & 29.28 & 98.49 & 0.73 & 20.42 & 0.12 & 2.94\\
     20 & \checkmark & \checkmark & & 90.65 & 248.87& 0.48& 15.08 & 0.35& 29.53& 103.46& 0.74& 20.74& 0.11&  2.40\\
  \bottomrule
\end{tabular}
}
\end{table*}

\begin{figure*}[h]
  \centering
  \includegraphics[width=\linewidth]{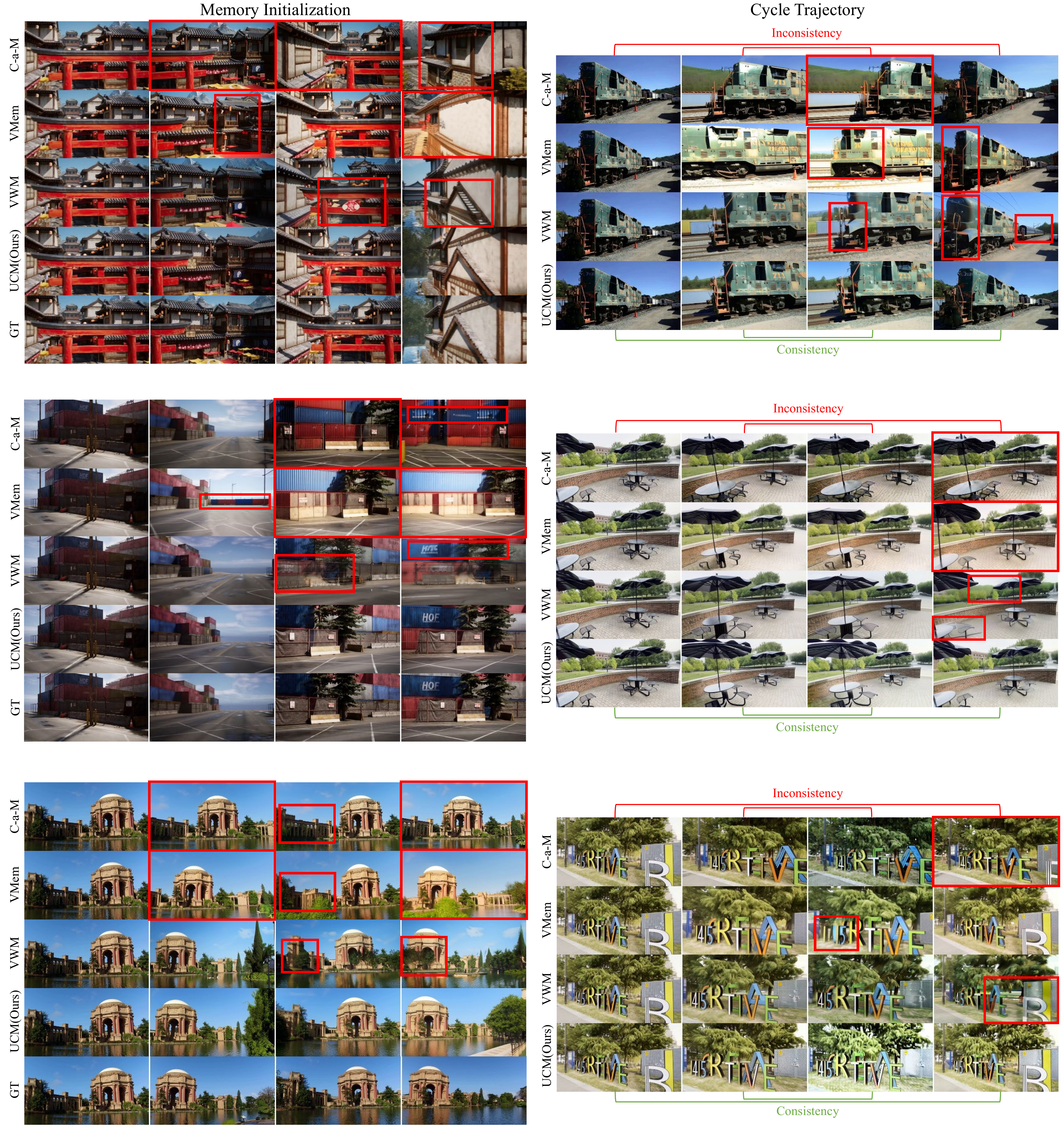}
  \caption{Supplementary visual comparison of long-term memory preservation. Red boxes indicate inaccurate camera control or scene inconsistency during generation.}
  \label{fig:supp_experiment}
\end{figure*}

As shown in Tab.~\ref{tab:mem}, our {\ourMethod} achieves the best performance under both evaluation settings, exhibiting significant improvements across all evaluation metrics. Qualitative comparisons are provided in Fig.~\ref{fig:main_experiment} and Fig.~\ref{fig:supp_experiment}. Implicit methods, such as C-a-M and VMem, lacking a token-level explicit correspondence prior, struggle to adhere faithfully to the camera trajectory  and sometimes fail to preserve long-term geometry (Fig.~\ref{fig:main_experiment}, left). Although VWM also demonstrates promising camera-controllability, it relies on TSDF fusion for aggregating multi-frame point clouds, leading to inflexibility for unbounded scenes (Fig.~\ref{fig:main_experiment}, left) or fine-grained structures (Fig.~\ref{fig:main_experiment}, right; Fig.~\ref{fig:supp_experiment}, right). In contrast, our method generates high-fidelity videos across both gaming and realistic scenarios, while achieving precise camera controllability and long-term scene consistency. We also provide more visual results in Fig.~\ref{fig:supp_vis}, which demonstrates the effectiveness of our proposed method on long-term scene-consistent world generation.

Tab.~\ref{tab:infer} presents the inference speed comparison across representative methods. For a fair evaluation, all approaches generate videos clip by clip with a total of 561 frames. Benefiting from our efficient dual-stream video diffusion design, our method achieves a higher inference efficiency than most competitors, except VWM. Unlike our approach, VWM adopts cross-attention trained from scratch, rather than extending the DiT attention sequence length, to capture inter-frame correspondence. Notably, we leverage STream3R~\cite{lan2025stream3r} to implement time-aware warped positional encoding derived from 3D reconstruction, where the inference time increases linearly with the KV cache length (we retain 1 frame out of every 4 frames in the KV cache). Advanced streaming reconstruction frameworks such as LoGeR~\cite{zhang2026loger} can further reduce this computational overhead and deliver additional speed gains.

\subsection{Ablation Studies}

\paragraph{Number of memory frames.} We explore how the number of retrieved historical frames affects memory capability in Tab.~\ref{tab:ablate}. As the number of retrieved frames increases, long-term memory preservation of {\ourMethod} improves under two settings with moderate computational overhead, facilitated by our proposed dual-stream video diffusion model. To balance the performance and computational cost, we retrieve 20 frames as our baseline for a good trade-off.

\paragraph{Efficient dual-stream video diffusion model.} To demonstrate the effectiveness of dual-stream architecture and the binary block mask in 3D attention, we ablate them by 1) concatenating both the noisy tokens and conditional tokens before feeding them into the diffusion model, or 2) applying the 3D full attention for injecting conditions. As shown in Tab.~\ref{tab:ablate}, although ablating the dual-stream design leads to an improvement under the cycle trajectory, it also significantly increases the computational cost and hinders practical applications. Notably, applying block attention masks not only accelerates the generation speed but also forces each token to attend its most relevant frame, resulting in an obvious performance gain.  

\paragraph{Data curation strategy.} To ablate data curation, we replace point cloud renderings with historical frames sampled from videos, leading to a consistent performance drop in terms of visual quality and recall consistency in Tab.~\ref{tab:ablate}. The comparison demonstrates that data curation allows our model to generate high-fidelity videos under diverse scenarios.

\begin{figure*}[h]
  \centering
  \includegraphics[width=\linewidth]{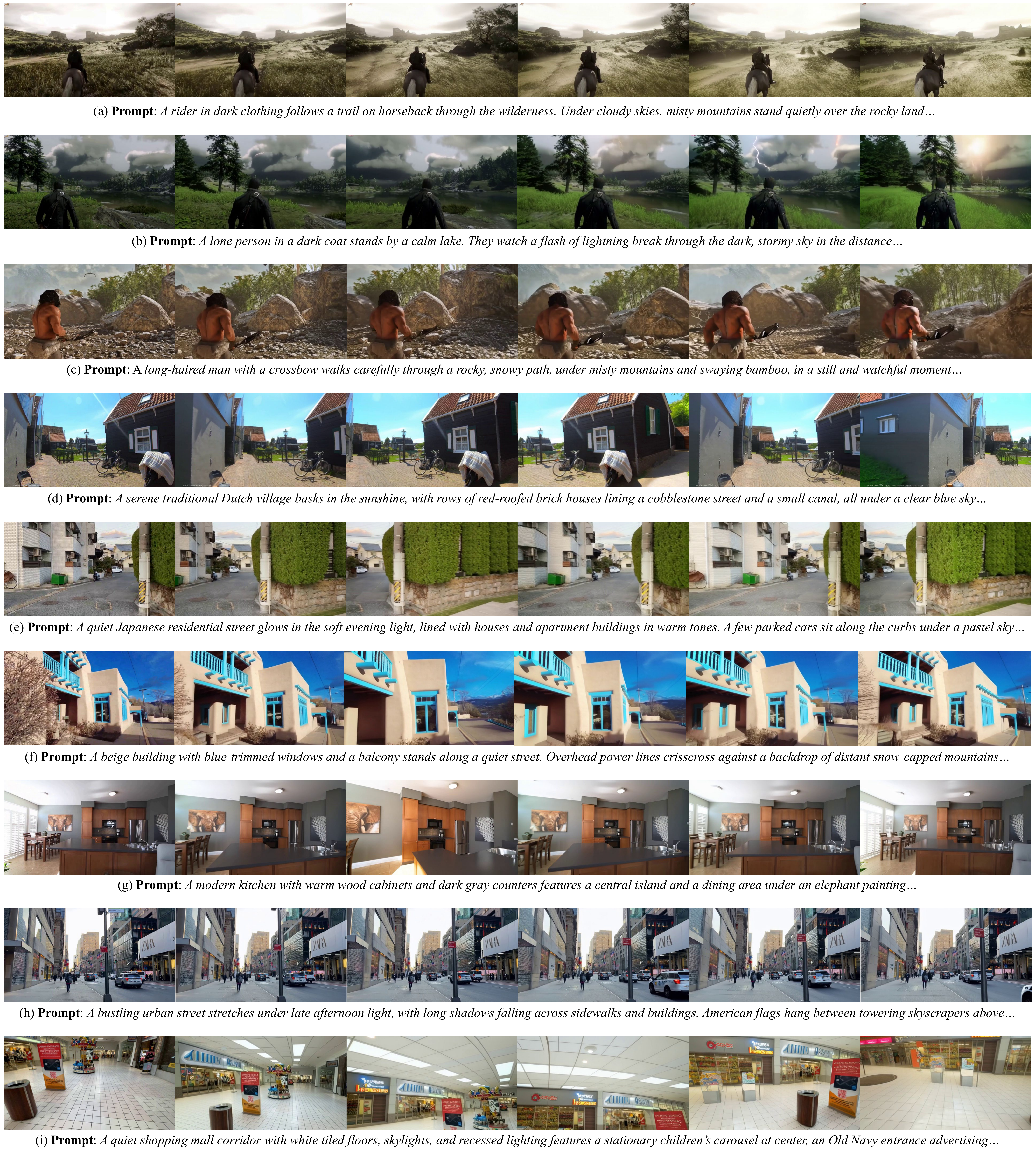}
  \caption{Supplementary visual results of our proposed {\ourMethod}. Starting from a reference image, {\ourMethod} can generate long-term videos that maintain scene consistency when viewing the same scene from different viewpoints.}
  \label{fig:supp_vis}
\end{figure*}

\paragraph{Limitations.} Although achieving promising generation, our proposed {\ourMethod} still suffers from the following limitations: 1) As shown in Fig.~\ref{fig:supp_vis} (a)(b), over clip-by-clip sequences, minor prediction errors accumulate, potentially impeding the appearance integrity of the simulation. 2) Our method relies on learned priors to distinguish dynamic objects and static scenes during memory injection, thus sometimes suffers from artifacts caused by movable objects. 3) As the number of generated frames increases, the storage and computational overhead of streaming depth estimation methods is non-negligible. How to efficiently organize historical information will be required for practical deployment. 

\section{Conclusion}

We present {\ourMethod}, a novel approach that realizes the unified modeling of camera control and memory via time-aware positional encoding warping for world models. To mitigate computational overhead during generation, we propose an efficient dual-stream video diffusion model, which incorporates block attention masks for memory and camera injection. Instead of relying on scarce long-term videos with
multiple revisits, we employ point cloud renderings to simulate revisiting, which enables us to exploit web-scale monocular videos to train our model. Extensive evaluations demonstrate that our method achieves high-fidelity video generation under precise camera control and long-term memory preservation, outperforming state-of-the-art approaches by a substantial margin.

\begin{acks}

Tian-Xing Xu, Zi-Xuan Wang and Zhongyi Zhang completed this work during their internship at Tongyi Lab, Alibaba. This work was supported by the National Key Research and Development Program of China (No. 2023YFF0905104), Fundamental and Interdisciplinary Disciplines Breakthrough Plan of the Ministry of Education of China (No. JYB2025XDXM101) and the Natural Science Foundation of China (No. 62132012, 62361146854).

\end{acks}

\clearpage
\bibliographystyle{ACM-Reference-Format}
\bibliography{sample-base}
\clearpage
\clearpage
\clearpage

\end{document}